\documentclass{article}

\usepackage{arxiv}

\usepackage[utf8]{inputenc} 
\usepackage[T1]{fontenc}    
\usepackage{hyperref}       
\usepackage{url}            
\usepackage{booktabs}       
\usepackage{amsfonts}       
\usepackage{nicefrac}       
\usepackage{microtype}      
\usepackage{lipsum}
\usepackage{graphicx}

\usepackage{tabularx}
\usepackage{enumerate}
\usepackage{amsmath}

\title{Triplet Permutation Method for Deep Learning of Single-Shot Person Re-Identification}

\author{
 M. J. Gómez-Silva, J.M. Armingol, A. de la Escalera\\
 Intelligent Systems Lab (LSI) Research Group, \\
 Universidad Carlos III de Madrid, \\ 
 Leganés, Madrid, Spain \\
 \texttt{magomezs@ing.uc3m.es}
}

\begin{document}
\maketitle
\begin{abstract}
 Solving Single-Shot Person Re-Identification (Re-Id) by training Deep Convolutional Neural Networks is a daunting challenge, due to the lack of training data, since only two images per person are available. This causes the overfitting of the models, leading to degenerated performance. This paper formulates the Triplet Permutation method to generate multiple training sets, from a certain re-id dataset. 
This is a novel strategy for feeding triplet networks,  which reduces the overfitting of the Single-Shot Re-Id model. The improved performance has been demonstrated over one of the most challenging Re-Id datasets, PRID2011, proving the effectiveness of the method.
\end{abstract}


\section{Introduction}
\label{intro}
Person Re-Identification (Re-Id) consists of recognising an individual across non-overlapping camera views, at different locations and time. It has attracted widespread attention in recent years due to its wide application in video surveillance tasks, such as threat detection, and multi-camera tracking.

Despite the considerable efforts of computer vision researchers focused on solving the Re-Id problem, it remains unsolved \cite{zheng2016person}.  This is due to the dramatic intra-class variations, caused by illumination, viewpoint and pose changes, which make pedestrians' appearance variates across camera views. Moreover, the presence of different people with similar appearance introduces inter-class ambiguities, turning the recognition of their identities into a daunting challenge. 

\begin{figure}[h]
  \centering
  \includegraphics[width=0.6\columnwidth]{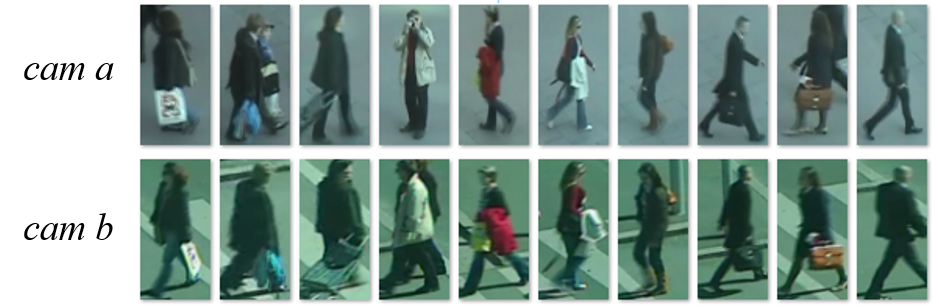}
\caption{Examples of matched pairs from PRID2011 dataset.} 
\label{fig:pairs}
\end{figure}

In order to solve this problem, earlier Re-Id systems generally fall into two categories: feature design \cite{matsukawa2016hierarchical, gomez2017deep}, and distance metric learning, \cite{weinberger2008fast,zheng2013reidentification} to compare pairs of person images. Most of the metric learning methods are applied over a set of previously computed features. Instead of that, this paper proposes the learning of a Deep Convolutional Neural Network (DCNN) to jointly model features and metric distance, for solving Single-Shot Re-Id, where only one image per person and per view is given. Fig.\ref{fig:pairs} shows examples of matched pairs (in each column), from PRID2011 dataset \cite{hirzer2011person}.

The first two works in Re-Id to use deep learning, \cite{li2014deepreid}, \cite{yi2014deep}, employed a Siamese model. This consists of two DCNNs, sharing parameters and joined in the last layer, where the loss function performs a pairwise verification. Therefore, the Re-Id task was treated as a pairwise binary classification to discriminate between matched and mismatched pairs of images.

Another learning model is the Triplet one, with three DCNN branches. It was presented in \cite{schroff2015facenet} where face recognition was addressed. 
This model receives three samples, two rendering the same person and the third one, a different identity, so it allows the comparison between a matched and a mismatched pair, and the objective function maximises the relative distance between them.
Some works, e.g. \cite{ding2015deep}, have extended the triplet model to the Re-Id problem. 
This paper proposes the use of the triplet model to train a VGG network \cite{simonyan2014very}. Other triplet approaches that also chose the VGG model to solve face recognition, \cite{zhuang2016fast}, and Multi-Shot Re-Id, \cite{liu2016end}.

However, Single-Shot Re-Id poses a unique challenge to deep supervised learning, since a highly layered model, with millions of parameters, must be trained on an extremely limited training set. Therefore, deep Re-Id models are prone to be overfitted and incapable to provide a general solution to identify unknown samples.
This paper refocuses the learning and feeding of a Re-Id triplet model by mean s of formulating a Triplet Permutation method, and analyses its effects over the learnt models. As result, of this study, this work contributes with the implementation of a novel triplets generation tool that allows increasing the variety of triplets from a certain re-id dataset (this is publicly available under http://github.com/magomezs/dataset\_factory).

The rest of the paper is organised as follows: Section \ref{architecture} presents the used learning model and Section \ref{permutation}, the proposed Triplet Permutation method to feed it. Section \ref{experimental_results} provides the experimental results, and some concluding remarks are given in Section \ref{conclusions}.

\section{Deep Re-Id Learning Model}
\label{architecture}

The objective in Single-Shot Re-Id task is to recognise the person appearing in an image from one view, called probe image, among all the images from the other view, called gallery images. With that purpose, the distance between the probe image and every gallery image is computed. Then, the gallery image presenting the shortest distance is selected as the correct match.

Instead of directly computing the distance over the raw images, these are calculated from representative feature arrays of the person images. Therefore, it is necessary to model an embedding $F(I)$, to map an input image $I$  to a feature space.
In this work, the feature embedding has been modelled by a DCNN, whose output gives the feature representation for an image, $F_{W}(I)$, so it depends on its weights values, $W$.

Choosing the number of layers of this DCNN is critical to get an effective Re-Id model. A high-layered network can embed salient and discriminative high-level features. Nevertheless, the increase in the number of layers of a model implies a larger number of parameters to train, boosting its overfitting. After several comparative experiments, an 11-layered network has been selected. Concretely, the A version of the set of Very Deep Convolutional Neural Networks presented in \cite{simonyan2014very}, hereafter called VGG11.
VGG11 presents eight convolution layers, three fully connected layers, and a SoftMax final layer, which has been removed in this work. Moreover, the input size has been modified to adapt it to the dimensions of the Re-Id samples.
Therefore, the input of the proposed DCNN is an RGB image of 128x64 pixels and its output is a point in the feature space represented by a 1000-dimensional array $F(I) \in R^n$ (n=1000). 
Besides, all hidden layers are provided of rectified linear units. Furthermore, as an extra measure to reduce overfitting, dropout has been implemented in the two first fully connected layers.

In order to train the network weights, a Triplet model has been chosen. The DCNN has been triplicated in three branches, which are forced to share the same weights. Therefore, the same feature embedding model is learned in each branch. Hence, every $i$ sample of the training dataset, $X$, is constituted by a triplet of person images, $X_i= \langle I_{i}^{a},I_{i}^{p},I_{i}^{n} \rangle $. The first branch of the triplet model receives an anchor sample, $I_i^a$, the second one takes an image, $I_i^p$, rendering the same person than the anchor image, and the input of the third branch is a different person's image, $I_i^n$. Therefore $I_i^a$ and $I_i^p$ form a matched pair of samples, hereinafter called a positive pair. On the contrary, $I_i^a$ and $I_i^n$ form a mismatched pair, called negative pair. 

The objective is to learn a transformation from the image to the feature space, such it leads the representations for the same person near, and far away from those of different people. This constraint is imposed by the triplet objective formulation, presented in \cite{schroff2015facenet}, and shown by Eq.\ref{eq:triplet}, which establishes a relative distance relationship. It requires the Euclidean distance for the negative pair to be larger than that for the positive one, by a predefined margin, $\tau$. This ensures that an individual's instance is closer to all samples of the same person than it is to any other people's instances, in the feature space.  
\begin{equation}
\label{eq:triplet}
\|F_W(I_s^a)-F_W(I_s^p)\|^2 + \tau <  \|F_W(I_s^a)-F_W(I_s^n)\|^2, \forall X_s\in X 
\end{equation}

This individual-meant approach results on clusters of samples of the same person. However in the Single-Shot Re-Id challenge, there is only a pair of images for each person, so the available data is not enough to adopt this approach. 
On the contrary, this work proposes to treat all the possible positive pairs as a set rendering the condition of similarity, and the negative ones, the dissimilarity situation. The discrimination between similarity and dissimilarity is learned by comparing every positive pair with all the possible negative pairs, taking advantage of the triplet model architecture for that. The training data has been generated by the method formulated in Section \ref{permutation}.

The learning process has been performed by the Triplet-based Mini-Batch Gradient Descent algorithm, presented in \cite{gomez2019balancing}. The chosen batch size (64) was that as large as the available memory resources make possible.


Furthermore, to alleviate the Re-Id data problem, it has been performed the transference of learning previously acquired from the Multi-Object Tracking domain to the Re-Id model, by means of initialising the model weights with the pre-trained values, as it is explained in \cite{gomez2019transferring}. Then, the Re-Id model is ready to be fine-tuned on the target Re-Id dataset, in order to acquire invariant and discriminative features.

\section{Triplet Permutation Method}
\label{permutation}

In the Single-Shot Re-Id challenge, the number of instances of a query identity is very limited in comparison with the vast quantity of potentially available different people representations. To overcome this limitation, the Triplet Permutation method generates all the possible triplet combinations from a certain Re-Id dataset. 
Three combining formulations have been designed, providing three different training triplets sets, $X_I$, $X_{II}$, $X_{III}$. They are presented by Eq. \ref{eq:setI}, \ref{eq:setII} and \ref{eq:setIII}, where $X_i=\left(I_i^a,{I}_i^p,{I}_i^n\ \right)$ represents every training triplet. Every image from the Re-Id dataset is denoted as $I_{id,c}$, where $id$ is the identity number, and $c$ is the camera-view from which it was captured. In Re-Id, two camera-view are involved, $A$ and $B$, so most of the rendered individuals, have two instances, $I_{id,A}$ and $I_{id,B}$. $M$ is the total number of training triplets.

\begin{equation}
\label{eq:setI}
     X_I:=\{  X_i=(I_i^a, I_i^p, I_i^n): I_i^a:= I_{j,x}, I_i^p:= I_{j,y}, I_i^n:= I_{k,y}  \Leftrightarrow j\neq k \wedge x=A \wedge y=B \wedge i \le M \}
\end{equation}

\begin{equation}
\label{eq:setII}
     X_{II}:=\{  X_i=(I_i^a, I_i^p, I_i^n): I_i^a:= I_{j,x}, I_i^p:= I_{j,y}, I_i^n:= I_{k,y}  \Leftrightarrow j\neq k \wedge x\neq y \wedge i \le M \}
\end{equation}

\begin{equation}
\label{eq:setIII}
     X_{III}:=\{  X_i=(I_i^a, I_i^p, I_i^n): I_i^a:= I_{j,x}, I_i^p:= I_{j,y}, I_i^n:= I_{k,z}  \Leftrightarrow j\neq k \wedge x\neq y \wedge i \le M \}
\end{equation}

By permuting, the camera set from which every branch takes its corresponding input image, the possible triplet arrangements are augmented. The first formulation is the simplest one, which takes the anchor samples from one view (A) and the positive and negative samples from the other one (B). Then, the second and third formulations increase the number of possible triplet combinations, as demonstrated below.

According to the first formulation, the number of possible positive pairs $(I_i^a,I_i^p)$ for each identity, $id$, is one, since there are only two views and the anchor is always taken from the same view. So, the number of total positive pairs is the number of identities, $P$.
In the second and the third cases, the number of positive pairs for each identity, $id$, are two, since the anchor can be taken from any of the two cameras. Although $(I_i^a:=I_{j,A}, I_i^p:=I_{j,B}) $ and $(I_i^a:=I_{j,B}, I_i^p:=I_{j,A})$ are identical pairs, they are deferentially considered. When adding a negative image, this is compared with the anchor images of the triplet. So, depending on which image of the pair is taken as an anchor, the triplet objective function will work differently, and for that reason, the pairs are considered as different ones.

Subsequently, as long as negative samples adding is concerned, in the first and the second formulations, the number of triplets generated from a positive pair is $(P-1)$. The reason is that the positive pair generated from an identity $id$ can be coupled with the remaining identities in the camera set from which the positive sample $I_i^p$ was taken. So, the total number of triplets generated in the first and the second cases are $P(P-1)$ and $2P(P-1)$, respectively.
The number of triplets generated from a positive pair by the third formulation is $2(P-1)$. That is because the positive pair generated from an identity $id$ can be coupled with the remaining identities in any of the camera sets. Therefore, the total number of triplets formed is $4P(P-1)$. 

In summary, the second formulation duplicates and the third one quadruplicates the number of training samples in comparison with the first one. Therefore, triplet augmentation is achieved through the proper combination of the samples. 

The Triplet Permutation method not only contributes to alleviating the model overfitting but also affects the features learning process. That is because feeding the triplet network with a particular training set, certain constraints are implicitly imposed in the learning process. These constraints are different for each set configuration, and they modify the characteristics of the solution in which the learning converges. 
The consideration of every training set formulation, Eq. \ref{eq:setI}, \ref{eq:setII} and \ref{eq:setIII}, together with the triplet objective, Eq. \ref{eq:triplet}, derives in the following deductions, where constraints 1, 2, 3, and 4 are denoted as $c_1$, $c_2$,  $c_3$ and $c_4$ and defined by Eq.\ref{eq:c1}, \ref{eq:c2}, \ref{eq:c3} and \ref{eq:c4}, respectively:

- Training set I, $X_I$, imposes $c_1$.

- Training set II, $X_{II}$, imposes $c_1$ and $c_2$.

- Training set III, $X_{III}$, imposes  $c_1$, $c_2$, $c_3$ and $c_4$.

\begin{equation}
\label{eq:c1}
       \forall j<P \land k<P: 
       \|F_W(I_{j,A})-F_W(I_{k,B})\|^2-\|F_W(I_{j,A})-F_W(I_{j,B})\|^2>\tau	 
\end{equation}

\begin{equation}
\label{eq:c2}
        \forall j<P \land k<P: 
        \|F_W(I_{j,B})-F_W(I_{k,A})\|^2-\|F_W(I_{j,B})-F_W(I_{j,A})\|^2>\tau
\end{equation}

\begin{equation}
\label{eq:c3}
        \forall j<P \land k<P: 
        \|F_W(I_{j,A})-F_W(I_{k,A})\|^2-\|F_W(I_{j,A})-F_W(I_{j,B})\|^2>\tau
\end{equation}

\begin{equation}
\label{eq:c4}
         \forall j<P \land k<P: 
         \|F_W(I_{j,B})-F_W(I_{k,B})\|^2-\|F_W(I_{j,B})-F_W(I_{j,A})\|^2>\tau
\end{equation}

\begin{figure}[h]
\includegraphics[width=\columnwidth]{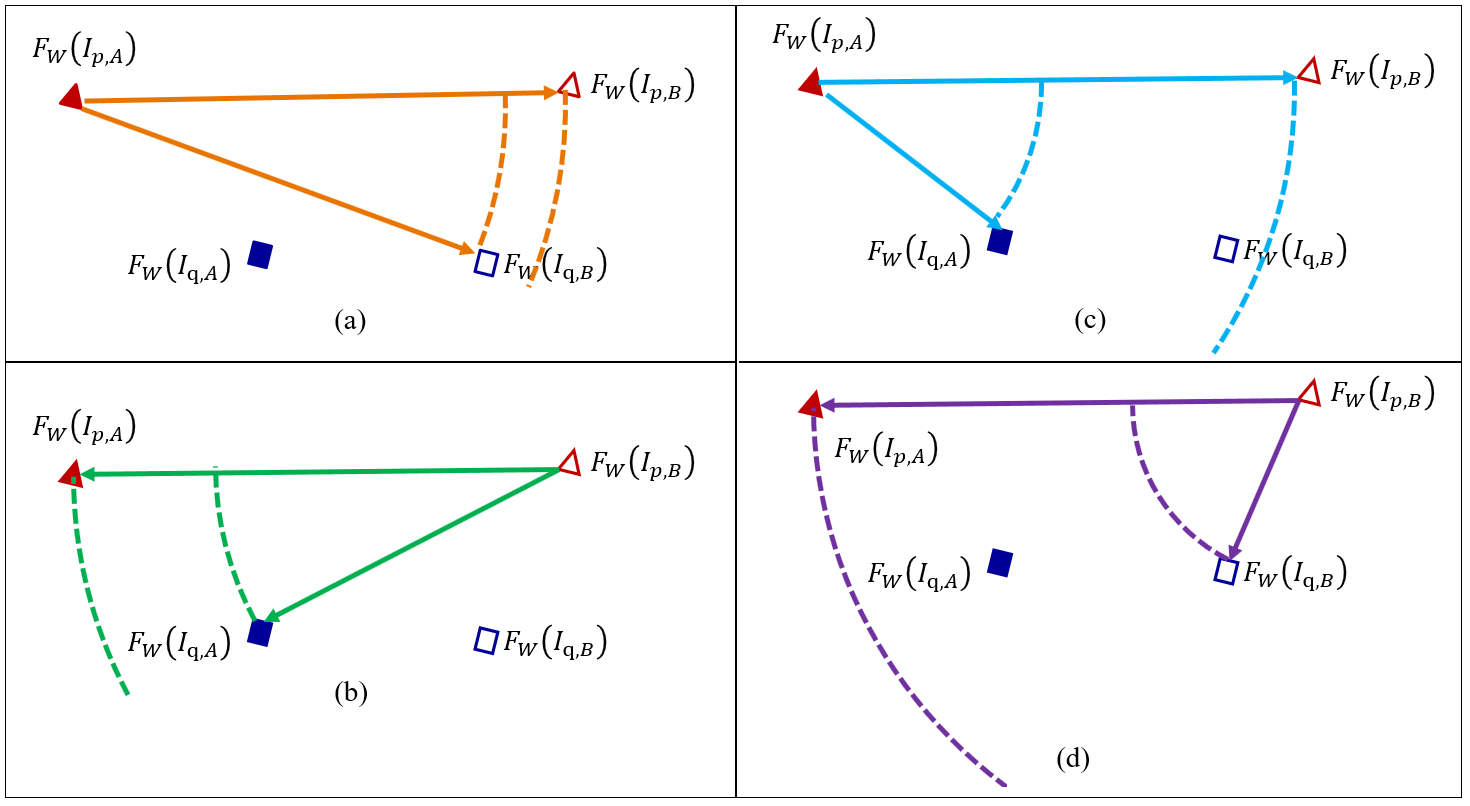}
\caption{Undesirable situations where the distance for positive pairs is larger than for negative ones, in the feature space.}
\label{fig:situations}
\end{figure}

During the training, the loss function leads the network weights to values that produce features able to meet the defined constraints, according to the selected training set. If eventually, these restrictions are complied by all the training samples, four undesirable situations (a, b, c, d) can be avoided. These situations, denoted as $s_a$, $s_b$, $s_c$ and $s_d$,  are described below, defined by Eq.\ref{eq:sa}, \ref{eq:sb}, \ref{eq:sc} and \ref{eq:sd}, respectively, and shown in Fig. \ref{fig:situations}.

\begin{enumerate}[a)]
    \item 	Sparsity of the positive pairs, with anchors in the cam-view $A$. The distance from an anchor image to its positive partner in camera $B$ is not smaller (in a $\tau$ measure) than the distance to any of the other samples from the view $B$.  
    \item Sparsity of the positive pairs, with anchors in the cam-view $B$. The distance from an anchor image to its positive partner in camera $A$ is not smaller (in a $\tau$ measure) than the distance to any of the other samples from the view $A$. 
    \item Clustering of different identities from the cam-view $A$. The distance from an anchor image from camera $A$ to some of the rest of samples from camera $A$ is not larger (in a $\tau$ measure) than the distance to its positive partner in the view $B$. 
    \item Clustering of different identities from the cam-view $B$. The distance from an anchor image from camera $B$ to some of the rest of samples from camera $B$ is not larger (in a $\tau$ measure) than the distance to its positive partner in the view $A$. 
\end{enumerate}

    \begin{equation}
        \label{eq:sa}
        \exists p<P\land q<P: 
        \|F_W(I_{p,A})-F_W(I_{q,B})\|^2-\|F_W(I_{p,A})-F_W(I_{p,B})\|^2<\tau
    \end{equation}
    
    \begin{equation}
        \label{eq:sb}
        \exists p<P\land q<P: 
        \|F_W(I_{p,B})-F_W(I_{q,A})\|^2-\|F_W(I_{p,B})-F_W(I_{p,A})\|^2<\tau
    \end{equation}   
    
    \begin{equation}
        \label{eq:sc}
        \exists p<P\land q<P: 
        \|F_W(I_{p,A})-F_W(I_{q,A})\|^2-\|F_W(I_{p,A})-F_W(I_{p,B})\|^2<\tau
    \end{equation}   
    
    \begin{equation}
        \label{eq:sd}
        \exists p<P\land q<P: 
        \|F_W(I_{p,B})-F_W(I_{q,B})\|^2-\|F_W(I_{p,B})-F_W(I_{p,A})\|^2<\tau
    \end{equation}  

The fact that different training sets configurations allows to avoid certain undesirable situations is demonstrated below. As an aside, it is necessary to clarify that the constraints $c_1$, $c_2$, $c_3$ and $c_4$ are imposed over all the training samples, so indexes $j$ and $k$ can take any value between one and the view sets sizes. Conversely, situations $s_a$, $s_b$, $s_c$ and $s_d$, might arise only for the particular identities rendered by indexes $p$ and $q$. 

\begin{figure}[h]
\includegraphics[width=\columnwidth]{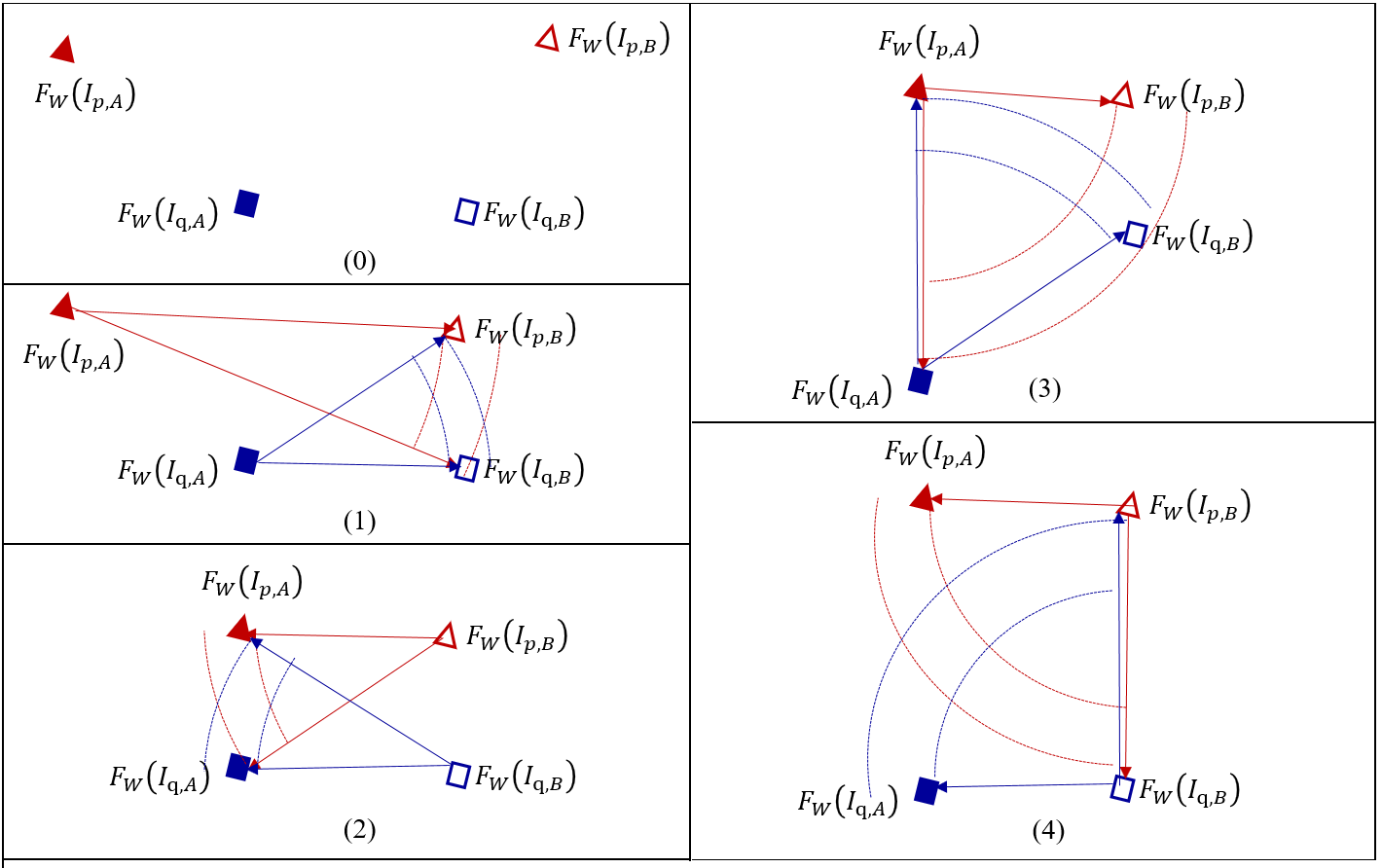}
\caption{Solving the undesirable situations $s_a$, $s_b$, $s_c$ and $s_d$ by imposing constraints $c_1$, $c_2$, $c_3$ and $c_4$, all rendered in the feature space. In (0) no constraint is imposed. In (1), (2), (3), (4) constraints $c_1$, $c_2$, $c_3$ and $c_4$ are consecutively added.}
\label{fig:constraints}
\end{figure}

If $(I_{p,A}, I_{p,B}, I_{q,B})$ is forced to meet the condition defined in Eq. \ref{eq:sa} (i.e. $j$ index takes identity $p$, and $k$ index takes identity $q$), the resulting proposition is just the opposite of the definition of $s_a$, as  Eq. \ref{eq:nosa} demonstrates. Hence, no triplet forced to meet $c_1$ can produce situation $s_a$, but the prevention of situations $s_b$, $s_c$ and $s_d$ is not ensured, as Fig. \ref{fig:constraints}(1) shows. 

If $(I_{p,B}, I_{p,A}, I_{q,A})$ is forced to meet the condition defined by Eq. \ref{eq:sb}, the resulting proposition is just the opposite of the definition of $s_b$, as  Eq. \ref{eq:nosb} demonstrates. Hence, no triplet forced to meet $c_2$ can produce situation $s_b$, but the prevention of situations $s_a$, $s_c$ and $s_d$ is not ensured, as Fig. \ref{fig:constraints}(2) shows. 

If $(I_{p,A}, I_{p,B}, I_{q,A})$ is forced to meet the condition defined by  Eq. \ref{eq:sc}, the resulting proposition is just the opposite of the definition of $s_c$, as  Eq. \ref{eq:nosc} demonstrates. Hence, no triplet forced to meet $c_3$ can produce situation $s_c$, but the prevention of situations $s_a$, $s_b$ and $s_d$ is not ensured, as Fig. \ref{fig:constraints}(3) shows.

If $(I_{p,B}, I_{p,A}, I_{q,B})$ is forced to meet the condition defined by  Eq. \ref{eq:sd}, the resulting proposition is just the opposite of the definition of $s_d$, as  Eq. \ref{eq:nosd} demonstrates. Hence, no triplet forced to meet $c_4$ can produce situation $s_d$, but the prevention of situations $s_a$, $s_b$ and $s_c$ is not ensured, as Fig. \ref{fig:constraints}(4) shows.

\begin{equation}
\label{eq:nosa}
      \|F_W(I_{p,A})-F_W(I_{q,B})\|^2 -\|F_W(I_{p,A})-F_W(I_{p,B})\|^2 >\tau =\neg s_a
\end{equation}

\begin{equation}
\label{eq:nosb}
   \|F_W(I_{p,B})-F_W(I_{q,A})\|^2-\|F_W(I_{p,B})-F_W(I_{p,A})\|^2 >\tau =\neg s_b	
\end{equation}

\begin{equation}
\label{eq:nosc}
   \|F_W(I_{p,A})-F_W(I_{q,A})\|^2- \|F_W(I_{p,A})-F_W(I_{p,B})\|^2 >\tau =\neg s_c
\end{equation}

\begin{equation}
\label{eq:nosd}
   \|F_W(I_{p,B})-F_W(I_{q,B})\|^2- \|F_W(I_{p,B})-F_W(I_{p,A})\|^2 >\tau =\neg s_d
\end{equation}

In conclusion, training set I avoid situation $s_a$, training set II, $s_a$ and $s_b$, and training set III, $s_a$, $s_b$, $s_c$ and $s_d$. Therefore, Triplet Permutation not only helps to prevent overfitting, but also the convergence of the network in a model that does not let situations $s_a$, $s_b$, $s_c$ and $s_d$ arise for the training samples.

\section{Experimental Results}
\label{experimental_results}
To evaluate the Triplet Permutation method, the VGG11 network has been trained on each one of the three generated training sets, resulting on three experiments, listed in Table \ref{tab:exp}.
\begin{table}[h]
\begin{center}
\caption{Experiments settings}
\label{tab:exp}   
\begin{tabularx}{0.3\columnwidth}{l l}
\hline
Experiment & Used training set \\
\hline
VGG11\_A &	Training set I. \\
VGG11\_B &	Training set II. \\
VGG11\_C &	Training set III. \\
\hline
\end{tabularx}
\end{center}
\end{table}

The Permutation method has been applied over samples from the PRID2011 dataset \cite{hirzer2011person}. This is a representative example of dataset acquired from a pair of fixed non-overlapping views, so this is composed of two sets of person images, each one containing the images acquired from one of the two views, with remarkable differences in camera characteristics, illumination, person's poses, and background. On contrary, other Re-Id datasets were captured from multiples views, such as VIPeR \cite{gray2008viewpoint} and CUHK-02 \cite{li2012human}, and they present images from different views within the same set of images (probe or gallery). 

On PRID2011 dataset, set A contains 385 images, and set B, 749. 200 of the captured individuals are rendered in both sets, and 100 of them were randomly selected as training samples, and they were properly combined by the Triplet Permutation method to generate a huge training set. The test set has been formed by following the procedure described in \cite{hirzer2011person}, i.e. the images of set A for the 100 remaining individuals with representation in both sets have been used as the probe set. The gallery set has been formed by 649 images belonging to set B (all images of set B except the 100 samples corresponding to the training and cross-validation individuals).

The learnt models are evaluated by their Cumulative Matching Characteristic (CMC) curve, which is a standard metric of Re-Id performance.
To obtain that, first, a probe image is coupled with every image from the gallery set and the squared Euclidean distance between their features is computed. The distance values are ranked, so that, the matches presenting the lowest distances are considered as the top matches. This process is repeated for every probe image. Then the CMC curve renders the expectation of finding the correct match within the top $r$ matches, for different values of $r$, called ranks. 

\begin{figure}[!h]
\includegraphics[width=\columnwidth]{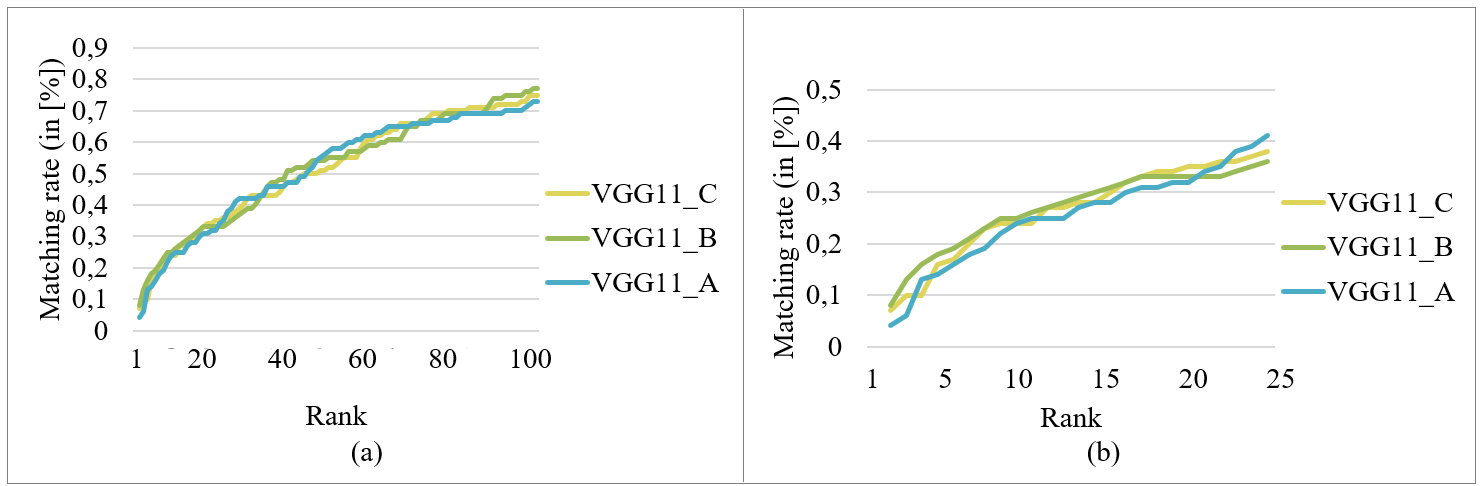}
\caption{CMC curves of models learnt on different triplets training sets from PRID2011 dataset, for the first 100 ranks in (a), and detailed for the first 25 ranks in (b).}  
\label{fig:cmc_sets_prid}
\end{figure}

Table \ref{tab:cmc_prid} presents the CMC scores of the models trained over the three proposed triplets sets, and Fig.\ref{fig:cmc_sets_prid}, their CMC curves, in ease of comparison. 
The best scores are highlighted in bold.

\begin{table}[!h]
\centering
\caption{CMC scores (in [\%]) for models learnt with different conducted experiments over  PRID2011 dataset}
\label{tab:cmc_prid}
\begin{tabular}{l l l l l l l}
\cline{2-7}
 & \multicolumn{6}{c}{Rank}\\
\cline{2-7}
 &	1	& 5& 	10& 	20&	50&	100\\
\hline
VGG11\_A	&4	&16	&25	&32	&\textbf{58}	&73\\
VGG11\_B	&\textbf{8}	&\textbf{19}	&\textbf{26}	&33	&55	&\textbf{77}\\
VGG11\_C	&7	&17	&24	&\textbf{35}	&53	&75\\
\hline
\end{tabular}
\end{table}
Experiment VGG11\_B (training set II), presents the highest scores, especially in the first ranks, since the number of triplets combination created is augmented with respect to training set I. This augmentation reduces the overfitting. However, although the number of triplets obtained in training set III is larger, this does not improve the algorithm performance. Set III contains triplets in which negative pairs are formed with images from the same view and this situation never happens in the test set. 

In conclusion, the permutation of training samples increases the model performance, as long as, the training conditions are similar to the test conditions. Otherwise, the model would have been trained for a task with different specifications.

The CMC scores of the model obtained by experiment VGG11\_B is compared with other Re-Id methods in Table \ref{tab:comparison_prid}.

\begin{table}[!h]
\centering
\caption{ Comparison of CMC rates (in [\%]) of re-id methods on PRID2011 dataset, ‘-’ indicates no result was reported.}
\label{tab:comparison_prid}       
\begin{tabularx}{8.5cm}{X l l l l l l}
\cline{2-7}
 & \multicolumn{6}{c}{Rank}\\
\cline{2-7}
 &	1	& 5& 	10& 	20&	50&	100\\
\hline
Proposed method	&\textbf{8}	&\textbf{19}	&\textbf{26}	&\textbf{33}	&\textbf{55}	&\textbf{77}\\
PSFI+PRDC \cite{liu2014evaluating}	&3	&9	&16	&24	&39	&-\\
PRDC \cite{zheng2013reidentification}&3	&10	&15	&23	&38	&-\\
PSFI+RankSVM \cite{liu2014evaluating}	&4	&9	&13	&20	&32	&-\\
RankSVM \cite{prosser2010person}	&4	&9	&13	&19	&32	&-\\
LDA \cite{fisher1936use}	&4	&-	&14	&21	&35	&48\\
GFI \cite{loy2010time}	&4	&-	&10	&17	&32	&-\\
Euclidean \cite{hirzer2012relaxed}	&3	&-	&10	&14	&28	&45\\
LDML \cite{guillaumin2009you}	&2	&-	&6	&11	&19	&32\\
PSFI \cite{liu2014evaluating}	&1	&2	&4	&7	&14	&-\\
\hline
\end{tabularx}
\end{table}

The recent trend on Re-Id research is the use of tracklets of the individuals to re-identify them, using Multi-Shot datasets, such as CUKH03 \cite{li2014deepreid} and Market1501 \cite{zheng2015scalable}. However, the objective of this work is to keep the Single-Shot approach to reduce the quantity of data transmitted between cameras, and speed up the communications in a distributed network of cooperative sensors. 

The main purpose of this paper is to alleviate overfitting, to make deep Re-id model performance overcomes methods based on the design of hand-crafted features and metric distance learning. This achievement is shown in Table \ref{tab:comparison_prid}, where an extensive list of method’s performances is presented. Indeed, this result can be taken as proof of concept to verify that, despite the data problem, deep learning can be exploited to solve Single-Shot Re-Id. Consequently, this also demonstrates that research on new overfitting reduction techniques, like Triplet Permutation, is worthy of being developed.

In \cite{hirzer2012relaxed}, the Euclidean distance is directly applied to compare hand-crafted person representations. Our model also uses the Euclidean distance, with the difference that the compared descriptors have been learnt by a deep Re-Id neural model. Other methods are focused on finding the proper combination of features to represent a person image, like Ranking Support Vector Machines (Rank-SVM), \cite{prosser2010person}. The features embedding is automatically found by our neural model, which produces a remarkable improvement in the performance.

Other works apply general metric learners, such as Probabilistic Relative Distance Comparison (PRDC), \cite{zheng2013reidentification}, Logistic Discriminant Metric Learning (LDML) \cite{guillaumin2009you} and Linear Discriminant Analysis (LDA) \cite{fisher1936use}. In general, these methods learn a metric distance over a set of previously computed features. On the contrary, this paper proposes a learning model to automatically and jointly find salient features and their proper combination, providing comparable or even better results.

In \cite{liu2014evaluating}, a Prototype-Sensitive Feature Importance based method is proposed to adaptively weight features according to different groups of the population, and the combination of this approach with previous methods (PSFI+PRCD and PSFI+RankSVM). On contrary, \cite{loy2010time} presented a Global Feature Importance (GFI) approach. No population discrimination has been made in the proposed method, and the general weighting of the features to create a global descriptor have been implicitly performed by the proposed deep model learning.

\section{Conclusions}
\label{conclusions}
In order to cope with the lack of labelled available data for Single-Shot Re-Identification, a method denoted as Triplet Permutation has been formulated. This allows generating a vast amount of triplet combinations from a certain Re-Id dataset, under different constraints, preventing extra data acquisition and labelling costs. Its main goal is to alleviate the overfitting from which deep re-identification models usually suffer. 

The application of the presented technique has been analysed and evaluated over one of the most challenging and commonly used re-id datasets: PRID2011 \cite{hirzer2011person}. The results have proved the effectiveness of the proposed techniques to improve the performance of the proposed re-id model, which outperforms some of the most used metric learning methods.  

This paper demonstrates the potential application of deep learning to solve the challenging task of Single-Shot Re-Id, as long as the lack of data is faced by a proper strategy.


\section{Acknowledgements}
This work was supported by the Spanish Government through the CICYT projects (TRA2015-63708-R and TRA2016-78886-C3-1-R), Ministerio de Educación, Cultura y Deporte para la Formación de Profesorado Universitario (FPU14/02143), and Comunidad de Madrid through SEGVAUTO-TRIES (S2013/MIT-2713). We are grateful to NVIDIA Corporation for the donation of the GPUs used for this research.

\bibliographystyle{unsrt}  
\bibliography{main}  

\end{document}